\documentclass[sigconf]{acmart}

\AtBeginDocument{%
  \providecommand\BibTeX{{%
    \normalfont B\kern-0.5em{\scshape i\kern-0.25em b}\kern-0.8em\TeX}}}

\setcopyright{acmcopyright}
\copyrightyear{2020}
\acmYear{2020}
\acmDOI{10.1145/3377929.3398075}

\acmConference[GECCO '20 Companion]{GECCO '20: ACM Genetic and Evolutionary
Computation Conference 2020}{July 8--12, 2020}{Canc\'un, Mexico}
\acmBooktitle{ACM Genetic and Evolutionary
Computation Conference}
\acmPrice{15.00}
\acmISBN{978-1-4503-7127-8/20/07}

\begin{document}
\title{Path Towards Multilevel Evolution of Robots}

\author{Shelvin Chand}
\affiliation{%
  \institution{Robotics and Autonomous Systems
Group, CSIRO}
  \streetaddress{1 Th{\o}rv{\"a}ld Circle}
  \city{Brisbane}
  \country{Australia}}
\email{shelvin.chand@csiro.au}

\author{David Howard}
\affiliation{%
  \institution{Robotics and Autonomous Systems
Group, CSIRO}
  \streetaddress{1 Th{\o}rv{\"a}ld Circle}
  \city{Brisbane}
  \country{Australia}}
\email{david.howard@csiro.au}

\renewcommand{\shortauthors}{Chand and Howard}

\begin{abstract}
 Multi-level evolution is a bottom-up robotic design paradigm which decomposes the design problem into layered sub-tasks that involve concurrent search for appropriate materials, component geometry and overall morphology. Each of the three layers operate with the goal of building a library of diverse candidate solutions which will be used either as building blocks for the layer above or provided to the decision maker for final use. In this paper we provide a theoretical discussion on the concepts and technologies that could potentially be used as building blocks for this framework.  
\end{abstract}




\maketitle

\section{Introduction}

Multi-level evolution (MLE) \cite{Howard2019} is a three-layer architecture that creates advanced evolved robots. Each layer deals with a different aspect of robot design, starting with lowest tier where materials are discovered, followed by the second layer where the components are created by considering a combination of materials on a given geometry, and finally leading to the third layer where components are combined into specific body plans (with controllers) to form complete robots.  In this paper we discuss research areas that are well-positioned to contribute to the successful implementation of the MLE architecture for robotic design. We look into wide variety of domains, listing outstanding potential challenges, while also providing recommendations on a suitable path forward.

\section{Enabling Technologies}
\subsection{Bi-Level Optimization}
A bi-level optimization problem can be defined as one in which there are two optimization tasks, one of which is nested within the other \cite{Sinha2018}. The nested task is known as the lower level (follower's) optimization problem while the outer optimization task is known as the upper level (leader's) optimization problem. The lower level problem acts as a constraint for the upper level problem. Each level has its own decision vectors, objectives and constraints. 

Concepts from bi-level research can be applied to the modelling of MLE. The robot layer can be seen as the upper level problem while the component and material layer can be seen as the nested sub-problems. This will result in a case where the robot layer will parameterize the lower layers based on the structure of the overall morphology and the type of environment which will then create a constraint on the search space for the component and material layers.  Elements of this strong coupling between layers is (a more computationally expensive) alternative to the standard implementation\cite{Howard2019} that offers the potential to provide direction to the combinatorial search.

\subsection{Quality Diversity}
Howard et al. \cite{Howard2019} envisioned MAP-Elites \cite{mouret2015} to be at the core of the MLE architecture where each layer would have one or more feature libraries representing materials, components or complete morphologies. For each solution $x$, MAP-Elites maintains an objective value $f(x)$ and a $N$-dimensional feature vector characterising the solution across the various features of interest. For this reason a feature or behavior function $b(x)$ also needs to be defined which computes $x$'s value across the $N$ feature dimensions. Map-Elites maintains a 'feature map' which is essentially a combination of cells which represent some discretisation of the various features. Map-Elites seems to be the ideal approach for maintaining a diverse library of solutions across the various layers within the MLE architecture.  Novelty search \cite{Lehman2008Novelty}, can also potentially be used depending on the type of problem and search landscape, as it more directly encourages diversity.

\subsection{Feature Selection}
Illumination algorithms such as MAP-Elites operate on a feature map which is usually divided into grids or niches with each grid / niche representation some combination/intersection of features. Feature selection is therefore an important issue to consider, as the size of the feature map determines the required search and computational budget.  For this reason it is essential that unnecessary features or overlapping features are eliminated from the feature map. 

Methods such as evolutionary feature selection\cite{Xue2016}, Principle Component Analysis or LASSO  may be useful in trimming the feature dimensions. It may be better to use approaches which are computationally efficient for online feature reduction as this would require less user involvement and not require additional a-priori experimentation. 

\subsection{Representation}
Researchers have explored many different forms of representation for evolving robotic components. These can roughly be categorised into direct and indirect representation. Direct representation (eg. 3D voxel grids\cite{Cheney2013} , bezier splines\cite{Collins2019}, etc) is typically integer or binary-based and exhibits a (relatively) low level of phenotypic complexity. Indirect representation (eg. CPPNs \cite{Cheney2013}) is much more scalable and thus offers a higher level of complexity. Indirect representation is also typically much more difficult to evolve because of the indirect relationship between genotype and phenotype. Some researchers have also attempted to evolve entire morphologies by using representations such as directed graphs\cite{Sims94}, L-Systems\cite{Hornby2001}, CPPNs, etc.

Each MLE layer is free to employ different forms of representation to maximize search capability. For example, the component layer could use CPPNs to represent component designs whereas the robot layer could employ directed graphs similar to the ones proposed by Sims \cite{Sims94} to represent the connections and interactions between the components. Even across the component layer it is possible to employ varying representations. For example it may be better to use bezier curves for certain components instead of CPPNs. Determining the type of representation across layers could be something that is decided on a-priori by the decision makers and domain experts or it could be determined dynamically during search, e.g. by seeding the population with solutions encoded using different forms of representation and letting the forces of evolution and selection pressure determine the best one.

\subsection{Fitness Function}
Within the MLE architecture, simulation will be needed to determine the quality of robotic components and entire morphologies at the component and robot layer, respectively. It may well be possible or even necessary to utilize a set of simulators as opposed to just one. For example, some simulators may be better for designing certain components due to simulation time, availability of more accurate or descriptive simulation environments, etc. In this way different body parts can or may be evolved using different simulation engines or even a combination of engines for added accuracy. This would of course require some level of tuning to ensure that certain simulators are not biased towards reporting higher or lower fitness for the same component. Using a hybrid approach \cite{Howard2019} of mating virtual and physical individuals may offer promise in crossing the reality gap.

Even when using only virtual simulations, one thing to strongly consider would be simulation time, which becomes significant when we consider that each of the components will be evolved using their own evolutionary run to build their respective feature maps. As such, we consider the use of surrogate models \cite{JIN2011} to help manage the computation load. 

\subsection{Including Materials}
The lowest layer in the MLE architecture deals with physical materials needed to construct the different components. The feature map here can be pre-loaded with materials from existing material libraries or it can be dynamically constructed by evolving new materials. Evolving new materials is a challenging task since the material properties space is high-dimensional and is affected by a myriad of processing properties, requiring specialist modelling tools. Physical experiments may well be necessary to gain deeper insight into the different features and then to subsequently build proper models and simulations.


\section{Conclusion}
To summarise: 1.) Modelling the overall problem as an extension of a bi-level formulation may be a good approach. This will create a natural hierarchical link between the various levels and also allow for upper-level constraints to be effectively integrated in the lower level searches and is feasible in-principle to combine with MAP-Elites. 2.) MAP-Elites is a strong choice for conducting search for materials, components and overall morphologies.  3.) Online feature reduction methods can help improve the computational efficiency of map-Elites by getting rid of unnecessary or overlapping feature dimensions. 4.) In terms of solution representation, it all depends on the layer and type of search. There is no `one size fits all' approach as different layers may require different representations. 5.) The choice of simulator is important as it should provide accurate and diverse forms of simulation options to ensure that the final product is robust. Surrogate modelling may be necessary to manage the computational load. 6.) To start off, the material layer can be pre-loaded with existing material libraries/models and later effort can be focused on evolving new materials. 
\bibliographystyle{ACM-Reference-Format}
\bibliography{ref}

\end{document}